\pdfoutput=1

\documentclass[11pt]{article}

\usepackage[]{acl}

\usepackage{times}
\usepackage{latexsym}
\usepackage{multirow}
\usepackage{amsmath}
\usepackage{amsfonts}
\usepackage{mathtools}
\usepackage{amssymb}
\usepackage{enumitem}
\usepackage{enumitem}
\usepackage{bbm}
\usepackage{mwe}
\usepackage{changepage}
\usepackage{soul}
\usepackage{xcolor}

\usepackage[T1]{fontenc}

\usepackage[utf8]{inputenc}

\usepackage{microtype}

\title{Data Augmentation for Improving Tail-traffic Robustness \\
in Skill-routing for Dialogue Systems}

\author{Ting-Wei Wu$^{1,2}$, Fatemeh Sheikholeslami$^2$, Mohammad Kachuee$^2$,  Jaeyoung Do$^2$, Sungjin Lee$^2$ \\
  $^1$Georgia Institute of Technology,
  $^2$Amazon Alexa AI, Seattle, WA \\
  \texttt{waynewu@gatech.edu} \\
  \texttt{\{shfateme, kachum, domjae, sungjinl\}@amazon.com} \\}

\begin{document}
\maketitle
\begin{abstract}
Large-scale conversational systems typically rely on a skill-routing component to route a user request to an appropriate skill and interpretation to serve the request.
In such system, the agent is responsible for serving thousands of skills and interpretations which create a long-tail distribution due to the natural frequency of requests. For example, the samples related to play music might be a thousand times more frequent than those asking for theatre show times.
Moreover, inputs used for ML-based skill routing are often a heterogeneous mix of strings, embedding vectors, categorical and scalar features which makes employing augmentation-based long-tail learning approaches challenging.
To improve the skill-routing robustness, we propose an augmentation of heterogeneous skill-routing data and training targeted for robust operation in long-tail data regimes. We explore a variety of conditional encoder-decoder generative frameworks to perturb original data fields and create synthetic training data. 
To demonstrate the effectiveness of the proposed method, we conduct extensive experiments using real-world data from a commercial conversational system. Based on the experiment results, the proposed approach improves more than 80\% (51 out of 63) of intents with less than 10K of traffic instances in the skill-routing replication task.
\end{abstract}

\section{Introduction}

Recent large-scale conversational systems such as Amazon Alexa, Apple Siri, Google Assistant, and Microsoft Cortana have shown great promise toward natural human-machine interactions \citep{Sarikaya17}. Such systems often involve multiple ML-based components to fulfill user requests.
Components such as Automated Speech Recognition (ASR) to transcribe the request, Natural Language Understanding (NLU) to assign a user's utterance to a set of potential interpretations i.e. domains, intent, and parse sentence entities.
Then, based on the NLU interpretations and other contextual signals (e.g. device type), a skill routing component is to select the best NLU interpretation and route the request to an appropriate skill. 


Self-learning based on customer satisfaction metrics is the state-of-the-art method for the skill routing problem. Typically, the skill routing problem is cast as a contextual bandit to optimize a reward signal generated by ML-based customer satisfaction estimators. While such self-learning approach is promising in terms of scalability, in a commercial system with thousands of skills/intents creating a long-tail distribution and considering the disparities in estimation quality for the customer satisfaction signals for different traffic segments, it is often challenging to maintain routing quality for entire traffic by solely relying on bandit learning objective. To address such issues, current self-learning methods rely on replication objectives to ensure policy robustness across off-policy bandit updates~\citep{kachuee-etal-2022-scalable,kachuee2022constrained}.

In this work, we attempt to enhance the robustness of skill routing systems by augmenting the low-appearance domain and intent data subsets. 
Typically, for a certain request, we are given a set of routing candidates represented as hypotheses each composed of an ASR transcribed text as well as other categorical data fields such as NLU interpretations, device type, device status, and the proposed skill. 
A model-based skill routing system is trained by replicating the ideal skill-routing decisions represented through a (large) training set of such requests and their contextual signals coupled with the correct corresponding skill for routing. 
However, in practice such datasets exhibit an imbalanced traffic between common requests and tail requests, leading to low replication accuracy and robustness in these domains. Therefore, we are interested in augmenting training data for such low-count segments. 


However augmenting such \emph{heterogeneous} data is non-trivial. Most natural language processing tasks with only text inputs focus on pure text augmentation by perturbing original texts at token spans \citep{Louvan2020SimpleDA, ye-etal-2021-efficient, cbert18} or entire sentences \citep{Einolghozati19, Chen2021GOLDIO}. Several works also leverage paraphrasing techniques \citep{Einolghozati19, cho-etal-2019-paraphrase, gao-etal-2020-paraphrase} to enhance the robustness of task-oriented dialog systems. 
However, manual paraphrasing dataset preparation with intense laboring is required especially for tail requests. 
Conditional generative approaches \citep{yoo18, peng20, Duan2020PretrainAP, Xia2020CGBERTCT, malandrakis-etal-2019-controlled, qiu-etal-2020-structured} instead provide flexible solutions of modeling text distribution that introduces variability yet preserves top-level semantics, which is ideal for labor-free data augmentation.

Nevertheless, modeling such distributions remains challenging and mostly unexplored in the research community, especially in the context of dialogue systems. In this paper, we explore the idea of generative data augmentation based on variational autoencoders (VAE) and transformer architectures to generate samples from the conditional distribution of skill routing hypothesis. 
The main contributions of this paper are as follows:
\vspace{-7pt}
\begin{enumerate}[leftmargin=*]
    \setlength\itemsep{-0.5em}
    \item Introducing a data augmentation framework for generating heterogeneous features available in conversational assistant systems.
    \item Enriching the training set and enhancing the robustness of skill routing models by leveraging the data augmenters applied on perturbed samples.
    \item Conducting extensive experiments using data from a real-world conversational system to demonstrate the impact of the proposed methods on various metrics for routing robustness and generation quality.
\end{enumerate}

\begin{figure}[t]
  \centering
  \includegraphics[width=1\linewidth]{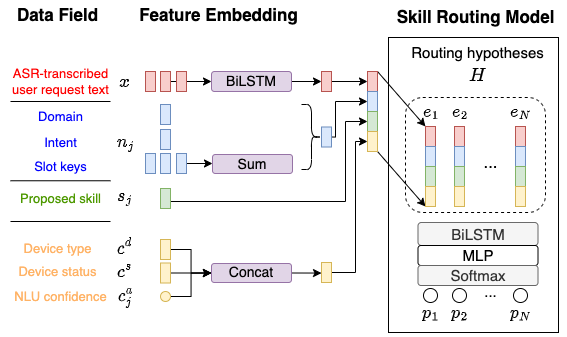}
  \caption{An overview of the base skill-routing system. 
  }
  \label{fig:base_model}
\end{figure}

\section{Problem Formulation}
\label{sec:problem}

\subsection{Skill routing}
\label{sec:skill_routing}

We consider the general skill routing problem in conversational systems. 
Given training dataset $\mathcal{D}=\{d_i\}$, each instance is represented as a pair $d_i = (H_i, a_i)$ with $H_i = \{ h_1, h_2, ..., h_N \}$ denoting a set of $N$ hypotheses and $a_i \in \{1,...,N\}$ denoting the logged action, i.e., the skill routing decision in the historical data, i.e., by the rule-based systems and/or previous ML model in production \citep{li21}.
The intermediary task to maintain policy robustness which is of focus here is to replicate the past actions $a_i$ of the deployed system by learning function $f_{\theta}(H_i)$ \citep{kachuee2021self}.

Let us define the replication accuracy for evaluating the robustness of $f_{\theta}$: 
\begin{align}\nonumber
    Acc(f_\theta, D_{test}) = 
    \sum_{(H_i, a_i) \in D_{test}} \frac{\mathbbm{1}[f_{\theta}(H_i)=a_i]}{|D_{test}|}
\end{align}
Thus, the offline evaluation metric $Acc(f_\theta, D_{test})$ captures how much $f_{\theta}$ matches the current stable routing behavior, which is considered a crucial element in hybrid policy setup \citep{kachuee-etal-2022-scalable}. 

\subsection{Heterogeneous data}
\label{sec:data_format}

A given hypothesis $h_j = (h^u_j, h^v_j)$ in the skill routing model input set $H_i$ contains utterance-level $h^u_j$ and hypothesis-level $h^v_j$ information, each of which has a collection of various data fields of different formats, leading to a heterogeneous dataset.
Utterance-level information $h^u_j$ are shared across all hypotheses in $H_i$ and includes the following data fields: (a) ASR-transcribed user request text $x$, (b) device type $c^d$, (c) device status $c^s$. Hypothesis-level information $h^v_j$ is instead unique to each hypothesis which has (d) NLU interpretation consisting of (domain,intent,slot) pairs $n_j$, (e) NLU confidence $c^a_j$, and (f) the proposed skill to serve the request $s_j$.
\section{Methodology}

\subsection{Base Model Architecture}
\label{sec:base_model}

\begin{figure}[t]
  \centering
  \includegraphics[width=1\linewidth]{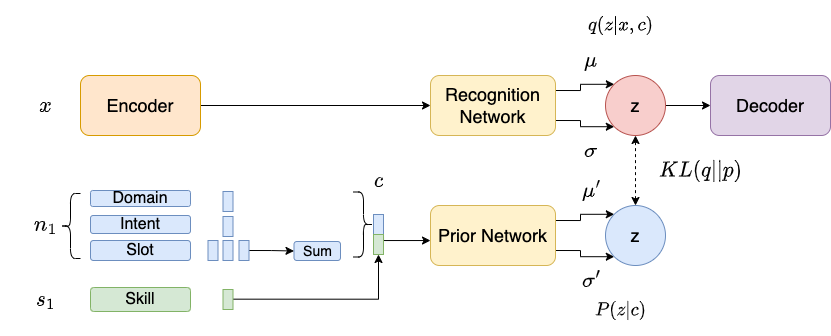}
  \caption{The schematic of the conditional VAE with prior network for text-only data augmentation.}
  \label{fig:vae_model}
\end{figure}

\begin{figure*}[t]
  \centering
  \includegraphics[width=1\linewidth]{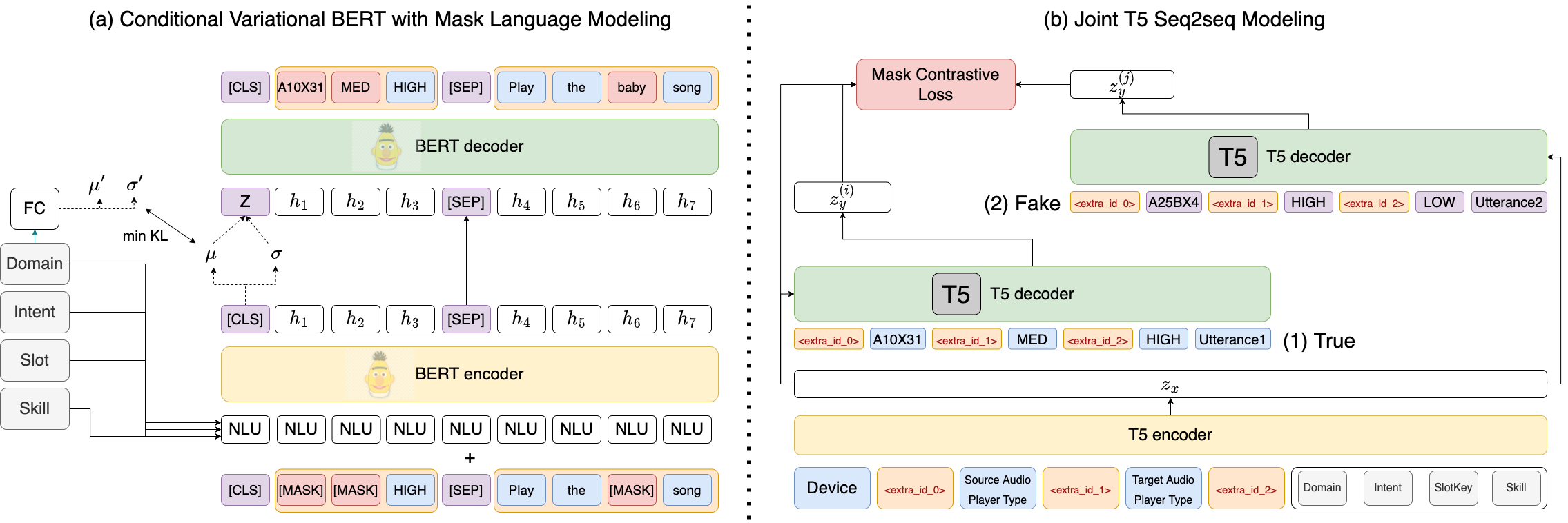}
  \caption{Two of main proposed generative frameworks for heterogeneous data augmentation. (a) Conditional Variational BERT structure by inputting a masked template and approximating the posterior distribution with variational inference. $h_1 \sim h_7$ are the hidden states at the middle of layers of BERT. (b) Joint T5 model structure with a sophisticated input-output sequence design, trained with mask contrastive learning and frequency-aware loss. The encoder states will be sent into the decoder with (1) true sequences and (2) false noises.}
  \label{fig:main}
\end{figure*}

Figure \ref{fig:base_model} shows our base model architecture, which for a given a list of $N$ hypotheses $H = \{ h_1, ..., h_N \}$, with $h_j=(h^u_j, h^v_j)$ defined as in Section \ref{sec:data_format}, proceeds as follows. First, the ASR-transcribed text $x$ is encoded using word vectors and a BiLSTM module. Words in other fields such as NLU interpretation entities are also  converted into fixed-size embedding vectors via the embedding modules, while all other categorical data fields $(c^d, c^s, n_j, s_j, c^a_j)$ represented as indices are encoded via trainable embedding layers (matrices). 
These feature vectors are then concatenated to get the final representation $e_j \in \mathcal{R}^d$ for hypothesis $h_j$. Embedded routing hypotheses $\{e_1, e_2, ..., e_N\}$ are then sorted based on the NLU interpretation confidence and are processed by a BiLSTM module followed by a MLP, finally yielding output action probabilities $[p_1,..., p_N]$.


\subsection{Heterogeneous Data Augmentation}
As discussed in the Section \ref{sec:skill_routing}, given the training set $\mathcal{D}=[\mathcal{D}_0;\mathcal{D}_T]$ comprised of frequent requests $\mathcal{D}_0$ and the tail requests $\mathcal{D}_T$,  we are interested in creating an augmentation set $\mathcal{D}_A = g_{\phi} (\mathcal{D}_T)$, which will then be incorporated during training  $\mathcal{D'} = [\mathcal{D}_0; \mathcal{D}_T; \mathcal{D}_A]$  to boost the replication accuracy of the low-traffic segments. Our proposition is to replace a subset of data fields described in Section \ref{sec:data_format} from $\mathcal{D}_T$ with the generated data from the proposed generative framework $g_{\phi}$, through implicitly modeling the joint data distribution of the heterogeneous feature sets $(h^u_j, h^v_j)$. 

We focus on augmenting the shared utterance-level fields $h^u$, including the ASR-transcribed text $x$, device type $c^d$ and device status signals $c^s$, with artificially generated ${h^u}'$. 
To do so, and in order to preserve the original semantics and prevent unrealistic deviation from natural data, 
We aim at modeling the joint data distribution $P({h^u}'|n_1, s_1)$, conditioned on $n_1$ the NLU interpretation (domain, intent, slot pairs) and the proposed skill $s_1$ from the top-1 hypothesis $h_1$ in a parametric way. 
The premise is that augmentation with different text and context variations can help 
boosting the routing robustness for such traffic segments. 


In the following sections, we propose three generative frameworks to generate such samples ${h^u}'$ to replace the original data fields $h^u$: 
(1) conditional VAEs with prior networks (pcVAE)
(2) conditional variational BERT with masked language modeling (MLM) (CV-BERT MLM), and 
(3) Joint T5 Seq2Seq modeling. 
The first two models are adaptations of the conditional variational auto-encoders, which the latter follows the BERT MLM task by modifying tokens in the original text. 
Since the T5 model is better capable of handling the heterogeneous nature of skill-routing data and shows higher quality of the generated text, and due to space limitation, the first two methods are discussed in Appendix \ref{sec:appendix_methods}, and the third model is discussed in section \ref{sec:t5}; see Figure \ref{fig:vae_model} and \ref{fig:main} for an overview.

\subsection{Joint T5 Seq2seq Model}
\label{sec:t5}



Conditional VAEs cannot produce high-quality texts with limited in-domain data and CV-BERT MLM aims to perturb the original text at the token level which preserves the structures yet sacrifices the diversity. 
Instead, we propose a transformer-based framework using T5 \citep{T520} that can more flexibly generate categorical fields and the utterance text from scratch.
T5 is an encoder-decoder model pre-trained on a multi-task mixture of unsupervised and supervised tasks and each task is converted into a text-to-text format. 
During pretraining, the spans of the input sequence are masked by the so-called sentinel tokens (a.k.a unique mask tokens) and the output sequence is formed as a concatenation of the same sentinel tokens and the real masked tokens. 
Herein, we also concatenate the categorical fields and top-1 NLU interpretation/skill $n_1, s_1$ and modify the inputs in Figure \ref{fig:main}. The intuition is to ask the T5 model to predict each categorical field one after another and accurately generate the utterance given $n_1, s_1$.

\noindent\textbf{Masked contrastive learning}: 
Vanilla T5 seq2seq models may suffer from \textit{exposure bias} problem where they are trained with teacher forcing and are never exposed to incorrectly generated tokens.
To ensure the fluency of generated sequences, we introduce a contrastive learning loss that explicitly forces the model to differentiate valid output sequences for a given input sentence.
We train the T5 model by maximizing the cosine similarity of 
positive representation pairs $(z_x^{(i)}, z_y^{(i)})$ and minimizing that of the negative pairs $(z_x^{(i)}, z_y^{(j)})$, which is randomly sampled from other non-target output sequence in the same batch.
\begin{align*}
    \mathcal{L}_{cont}(\theta)=\sum_{i=1}^N \log \frac{\exp(\langle z_x^{(i)},z_y^{(i)}\rangle/\tau)}{\sum_{z_y^{(j)} \in S} \exp(\langle z_x^{(i)},z_y^{(j)}\rangle/\tau)}
\end{align*}

In contrast with text summarization or translation tasks, we may encounter samples with different output sequences but the same input condition sequence $z_x^{(j)} = z_x^{(i)}$. It is not reasonable to consider these $z_y^{(j)}$ as negative samples to be contrasted. Instead, we apply a mask to only consider the negative samples in the same batch that have different inputs $z_x^{(j)} \neq z_x^{(i)} \in S$.

\noindent\textbf{Frequency-aware cross entropy loss}:
Maximum a-Posterior (MAP) objective may prompt the model to sample frequent and saver tokens, which is the last thing data augmentation would favor.
To avert the low diversity problem, we follow \citet{Jiang_2019} to introduce a Frequency-Aware Cross Entropy (FACE) loss by incorporating a weighting mechanism conditioned on token frequency. 
FACE loss function of predicted output token $y_t$ at time step $t$ could be expressed as:
\begin{align*}
    FACE(y_t) = -\sum^K_{i=1} w_i \delta_i(y_t) \log(P(c_i|y_{<t},X))
\end{align*}
where $K$ is the vocabulary size, $c_i$ is the candidate token, $\delta_i(y_t)=1$ if $y_t=c_i$ and $w_i$ will be calculated based on \textit{Pre-weight} and \textit{Post-weight} conditions. 
For \textit{Pre-weight}, $w_i=1 - \frac{f_i}{\max_j f_j}, \forall j \in \{1...K\}$ where $f_i$ is the relative frequency of the candidate token $c_i$. This function will penalize high frequency tokens to have lower weights. 
For \textit{Post-weight}, the function will try to penalize the model’s conservativeness: if the predicted token $y_t$ has a higher frequency than the ground truth $c_i$, which indicates the model conservatively picked a ``safe'' token, then its loss will be scaled up by $w_i$ > 1.
\begin{align*}
    w_i = 1+ \frac{ReLU(freq(y_t)-freq(c_i))}{\sum_j^K freq(c_j)}
\end{align*}



\begin{table*}[th]
\footnotesize
\centering 
\resizebox{0.9\linewidth}{!}{
\begin{tabular}{l|c|c|c||c||c|c|c|c||c|c|c}
\hline
\multicolumn{4}{c||}{\textbf{Model}} & \textbf{Fluency} & \multicolumn{4}{c||}{\textbf{Diversity}} & \multicolumn{3}{c}{\textbf{Field KL}} \\
\hline
\textbf{Base} & \textbf{MCL} & \textbf{PreW} & \textbf{PostW} & \textbf{Perplexity} & \textbf{Unique rate} & \textbf{Dist-1} & \textbf{Dist-2} & \textbf{Ent-4} & \textbf{Device} & \textbf{Source} & \textbf{Target} \\
\hline\hline
cVAE	    & - & - & - &	4.866	&	0.9027	&	0.086	&	0.553	&	9.946	&	-	&	-	&	-	\\
pcVAE	& - & - & - &	3.354	&	0.8984	&	0.078	&	0.492	&	9.878	&	-	&	-	&	-	\\
CV-BERT MLM	& - & - & - &	-	    &	0.6734	&	0.079	&	0.324	&	9.481	&	3.74E-02	&	4.80E-03	&	4.70E-03	\\
\hline
Joint T5 &  &  &  &	2.675	&	0.7893	&	0.057	&	0.280	&	9.538	&	5.85E-02	&	3.95E-03	&	4.22E-03	\\
Joint T5 & \checkmark &  &  &	2.653	&	0.8046	&	0.059	&	0.280	&	9.613	&	8.80E-02	&	1.72E-03	&	2.13E-03	\\
Joint T5 &  & \checkmark &  &	2.999	&	0.7375	&	0.050	&	0.262	&	9.393	&	7.63E-01	&	4.10E-03	&	4.44E-03	\\
Joint T5 &  &  & \checkmark &	\textbf{2.531}	&	\textbf{0.8065}	&	\textbf{0.063}	&	\textbf{0.293}	&	9.608	&	4.63E-02	&	1.56E-03	&	1.77E-03	\\
Joint T5 & \checkmark &  & \checkmark &	2.553	&	0.8007	&	0.060	&	0.291	&	\textbf{9.641}	&	\textbf{4.28E-02}	&	\textbf{1.51E-03}	&	\textbf{1.76E-03}	\\
\hline\hline
\multicolumn{12}{c}{\textbf{Test on low-count intents only}} \\
\hline
cVAE	    & - & - & - &	8.396	&	0.9591	&	0.259	&	0.795	&	7.739	&	-	&	-	&	-	\\
pcVAE	& - & - & - &	4.009	&	0.9332	&	0.275	&	0.843	&	6.897	&	-	&	-	&	-	\\
CV-BERT MLM	& - & - & - &	-	&	0.5886	&	0.213	&	0.638	&	7.218	&	1.26E-01	&	6.26E-03	&	5.78E-03	\\
\hline
Joint T5 &  &  &  &	2.543	&	0.6410	&	0.195	&	\textbf{0.648}	&	7.210	&	1.0454	&	3.84E-03	&	3.93E-03	\\
Joint T5 & \checkmark &  &  &	2.841	&	\textbf{0.6894}	&	0.187	&	0.647	&	\textbf{7.404}	&	3.1899	&	3.10E-01	&	3.16E-03	\\
Joint T5 &  & \checkmark &  &	2.935	&	0.6336	&	0.178	&	0.632	&	7.060	&	6.0623	&	5.54E-02	&	5.50E+02	\\
Joint T5 &  &  & \checkmark &	\textbf{2.307}	&	0.6520	&	0.211	&	0.642	&	7.127	&	\textbf{2.26E-01}	&	\textbf{2.47E-03}	&	\textbf{2.92E-03}	\\
Joint T5 & \checkmark &  & \checkmark &	2.325	&	0.6340	&	\textbf{0.218}	&	0.646	&	6.971	&	2.45E-01	&	5.52E-03	&	5.64E-03	\\
\hline

\end{tabular}
}
\caption{Intrinsic evaluation for all generation methods. We break down different versions of Joint T5 for comparison including: \textbf{MCL} with masked contrastive learning, frequency-aware loss with pre/post weight calculation (\textbf{PreW/PostW}). Second half of the table indicates the results of the test set on samples with low-count intents.} 
\label{table:intrinsic} 
\end{table*}
\section{Experiment Setting}

\subsection{Dataset}

We use data from a commercial conversational system in our experiments. 
We select 4 critical domains: \textit{Knowledge, Shopping, Video, Books} that have lower replication accuracy on the current skill routing production system. All data have been processed so that users are not identifiable (de-identified).
For training and intrinsic evaluation of the proposed generative models, we use a dataset of size 4M in domains of interest for training and 10K for validation. For testing, we prepare two test sets with one as 10k samples covering all 4 interested domains and another as 1k samples of selected low-count intents from such domains.
For the base skill routing model, the dataset contains another $\sim$ 80M traffic covering all 37 domains. We first filter the data with only 4 interested domains yielding ~5M data. We use the proposed data augmentation approaches to augment the original 5M data with 1:1 and 1:5 ratio. That is, each data point in this set is use to create 1 or 5 artificially generated samples by using the proposed models to re-generate one or more of the data fields while keeping the remaining fields intact. 
These 1X or 5X augmentation datasets are used with  the original 80M dataset to train the downstream skill-routing model; See App. \ref{sec:app_experiment_setup} for  setup details and network architecture.

\subsection{Evaluation metrics}


\noindent\textbf{Intrinsic evaluation}: These metrics evaluate the quality of the generated texts, irrespective of the influence on the downstream tasks: 
    \textbf{(a) Reconstruction accuracy}: we use test reconstruction loss, BLEU-1,2 between generated and original texts.
    \textbf{(b) Fluency}:  uses language model perplexity to evaluate how fluent the generated texts are.
    \textbf{(c) Diversity}:  uses the unique rate as the proportion of unique sentences across all sentences.
    \textbf{(d) Dist-1 and Dist-2}: for a more granular  evaluation of diversity, it calculates the percentage of the number of unique 1-grams/2-grams over all 1-grams/2-grams of tested sentence.
    \textbf{(e) Ent-4}: measures how evenly the empirical 4-gram distribution is for a given sentence.
    \textbf{(f) Field KL}: calculates the KL divergence between the original and the reconstructed categorical fields; see Appendix \ref{sec:app_intrinsic_metric} for more details.


\noindent\textbf{Extrinsic evaluation}:
We train skill routing models with the original training data and augmented data from different data augmentation methods, and evaluate the replication accuracy for each intent while focusing the report on the four domains of interest. Normal training (without any augmentation) serves as the baseline while results for training with 5X oversampling of the 4-domains is also report for benchmarking. 
In models trained with data augmentation, models are trained to jointly model the three fields $(x, c^d, c^s)$ defined in \ref{sec:data_format}, and in models denoted by "nlubin", we additionally change the nlu confidence field in each hypothesis with a probability of $0.8$, over values (HIGH, MEDIUM, LOW). The trained augmentation models are cVAE, pcVAE, conditional variational BERT with MLM training, and joint T5 seq2seq model. 

\section{Main Results}

\subsection{Intrinsic Evaluation}
Table \ref{table:intrinsic} presents different model comparison for the intrinsic metrics (see Appendix \ref{sec:app_intrinsic_eval} for additional comparisons).
VAE-based methods generate samples with higher diversity but with higher perplexity, while
CV-BERT MLM- which only modifies few tokens- shows a lower unique rate but better text quality. Overall,
Joint T5 models exhibit better text quality and fluency than VAE-based models, but with lower diversity since variations most derive from the token sampling at the output.
We found T5 with the masking contrastive loss has better diversity and T5 with post-weight frequency loss has the best performance both on perplexity and field generation.

{\textbf{Qualitative analysis}}: Table \ref{table:qualitative} gives  some generated text samples for the proposed generative models. It is observed that cVAE generates shorter colloquial texts while pcVAE introduces more interesting utterances yet with some grammatical errors; whereas T5 model generates more realistic text examples. Finally, 
CV-BERT MLM is able to replace some missing tokens with intriguing word fragments that preserve the original NLU interpretation. 
However, there are also  cases when the 
generated words are repeated with lower quality and may have syntax errors, e.g. ``what is the temperature in the center of the center of the earth.''


\begin{figure}[t]
  \centering
  \includegraphics[width=0.93\linewidth]{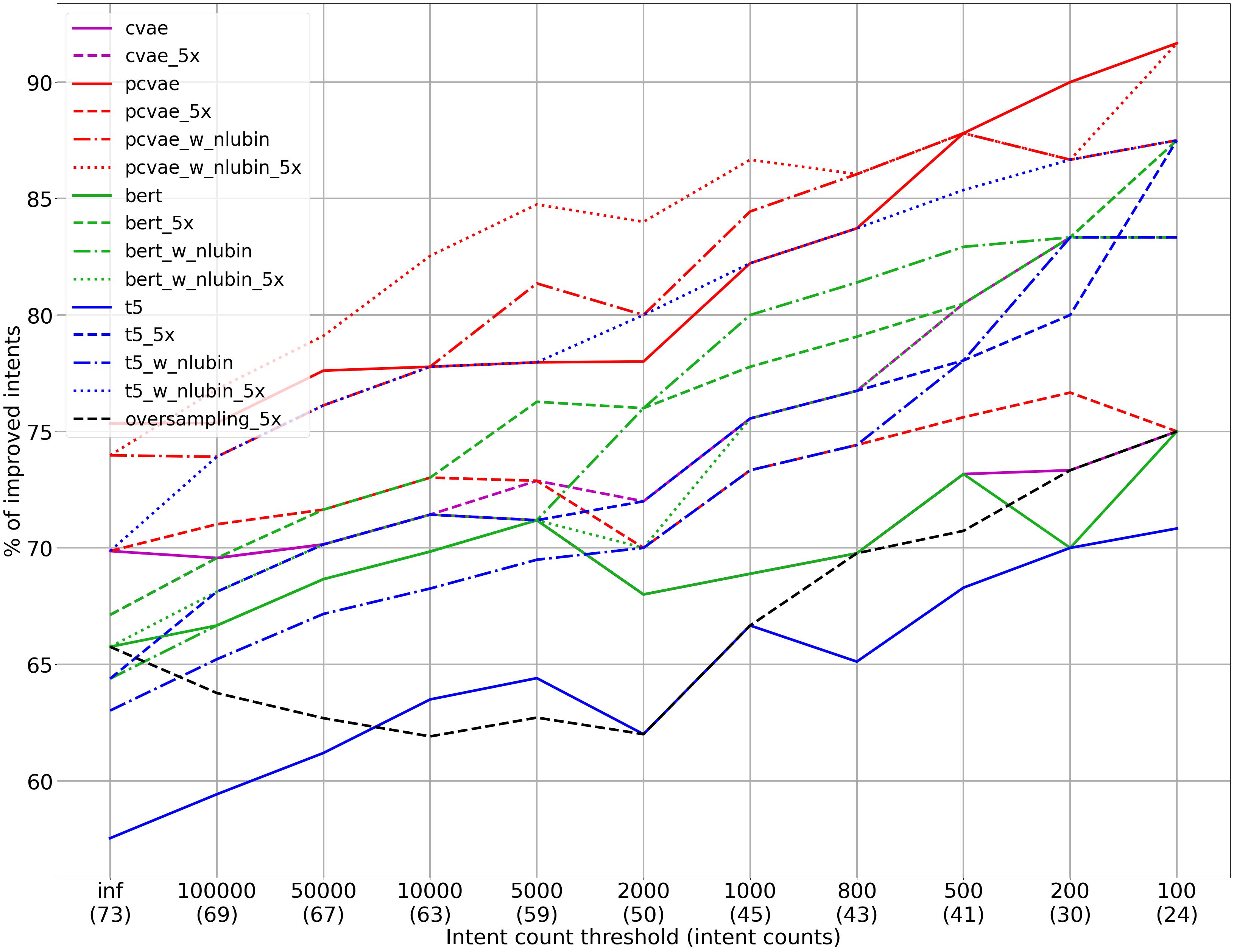}
  \caption{Percentage of intents that have higher accuracy than the baseline vs. intent count threshold. We show the total intent counts falling in each threshold category in the parentheses, and group similar models in the same color family for better illustration.}
  \label{fig:replication_improve}
\end{figure}

\begin{figure}[t]
  \centering
  \includegraphics[width=0.93\linewidth]{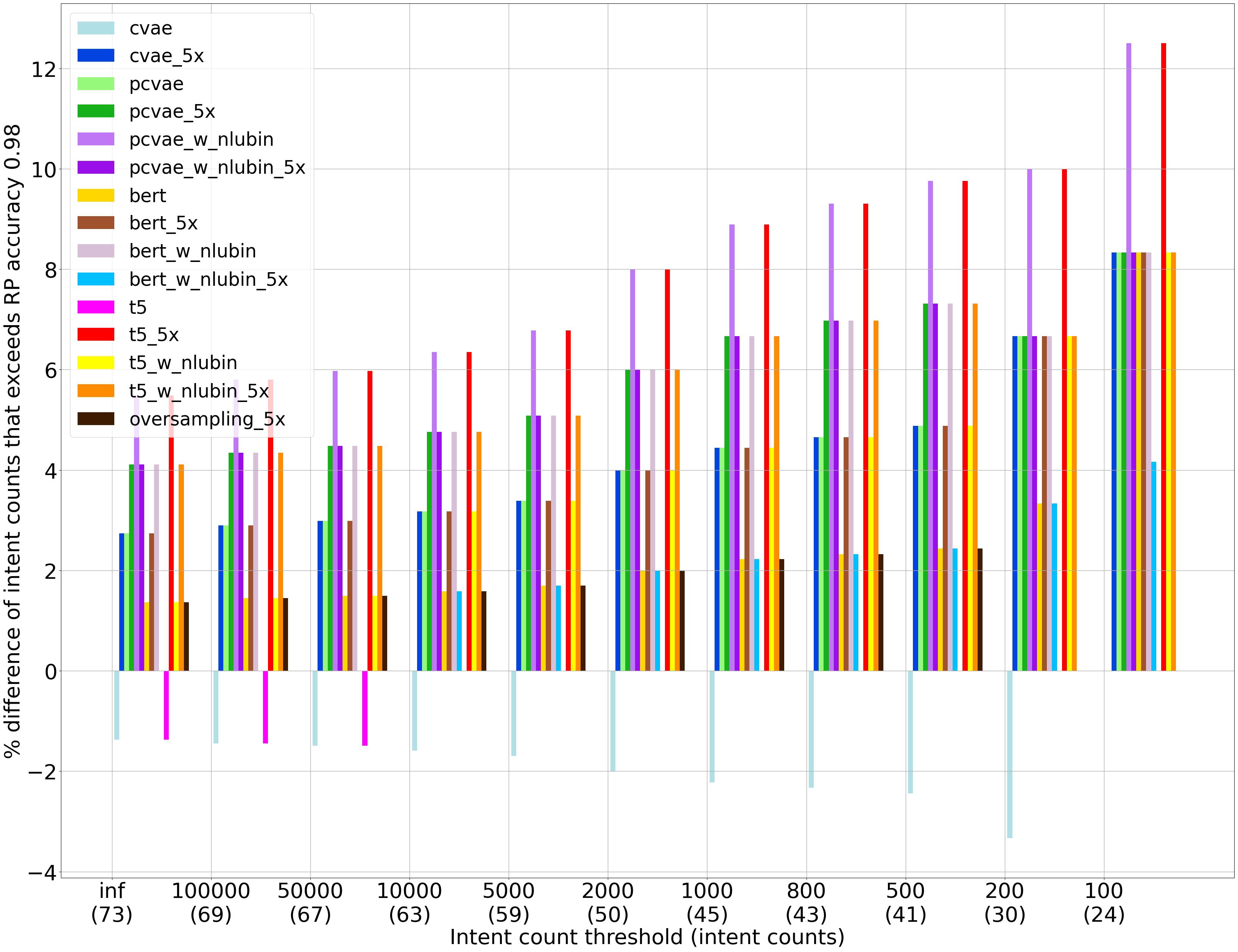}
  \caption{Percentage of intent count difference that exceeds 98\% replication accuracy vs. intent count threshold. Positive bars indicate the model has more intents with accuracy exceed 98\% than the baseline.}
  \label{fig:replication_improve_threshold}
\end{figure}

\begin{table}[t]
\footnotesize
\centering 
\resizebox{1.0\linewidth}{!}{
\begin{tabular}{l|l}
\hline
\textbf{NLU Interpretations} & \textbf{Domain: Knowledge, Intent: QA Intent, slotKey: Question}\\
\hline\hline
\textbf{Model} & \textbf{Generated texts}\\
\hline
\multirow{4}{*}{\textbf{cVAE}} & when is easter \\
                               & is all puma large \\
                               & what's super bowl i goes \\
                               & spell onomatopoeia \\
\hline
\multirow{4}{*}{\textbf{pcVAE}} & what is sixty four minus twenty \\
                               & how do you pronounce a word cable \\
                               & how many hours till the philadelphia world kansas basketball game tonight \\
                               & how do you probably is new york and fiftieth day\\
\hline
\multirow{5}{*}{\textbf{Joint T5}} & how old is mary gomez \\
                               & are there any more games tomorrow \\
                               & what's the current price of gold in the united states \\
                               & is it still good to plant a salmon \\
                               & what's the score of the nets play today \\
\hline
\multirow{6}{*}{\textbf{Joint T5 }} & what is the song called with the lyrics that she's not really gonna win the world \\
                               & what's the current time in Phoenix Arizona \\
                               & what's the most famous song in the world \\
                               & how long does it take a cat to eat \\
\textbf{(MCL \& PostW)} 
                                & tell me about this product of ninja turtles \\
                               & what's the movie with Kevin Durant in it \\
\hline
\multirow{5}{*}{\textbf{CV-BERT }} & who is \textbf{this spy} $\leftarrow$ \text{who is Sharon Carmichael}  \\
                                   & how many \textbf{syllables} does \textbf{April} have $\leftarrow$ \text{how many provinces does Canada have} \\
                               & what's \textbf{twenty hundred} divided by two $\leftarrow$ \text{what's one fourteen divided by two} \\
\textbf{MLM} 
                                & do you \textbf{know} about the\textbf{fire} discovery in\textbf{ Sacramento} California \\ 
                                & $\leftarrow$ \text{do you have information about the dinosaur discovery in Jamestown California} \\
                               & what is the \textbf{temperature} in the \textbf{center} of the \textbf{center} of the \textbf{earth} $\leftarrow$ \text{what is the} \\ & \text{ hole in the center of the iris of the eye} \\
\hline
\end{tabular}
}
\caption{Qualitative examples of generated texts from different generative models. Texts on the right of $\leftarrow$ in CV-BERT MLM are the original text. 
} 
\label{table:qualitative} 
\end{table}

\subsection{Extrinsic Evaluation}

To study the impact of different data augmentation techniques,  we utilize the augmentation sets in training the downstream skill-routing system and evaluate replication accuracy.
In Figure \ref{fig:replication_improve}, we calculate the percentage of intents that improves on replication accuracy, compared with baseline system (trained without augmentation). The figure is obtained by setting an intent count threshold $\tau$, keeping the intents whose count falls below $\tau$, and reporting the percentage of intent improvements by varying $\tau$. All proposed data augmentation approaches show improvements across different thresholds, where lower-count intents have larger improvement as that augmented data better helps these tail segments, while simply duplicating data (oversampling) helps with the performance to a lesser degree. Surprisingly, some VAE models with higher diversity yet poorer quality lead to better improvements, however increasing augmentation size is helpful in high-quality texts by T5 model. 

In Figure \ref{fig:replication_improve_threshold}, for each method, we calculate the percentage of  intent with  replication accuracy exceeding 98\%. After subtracting the percentage of intents in the baseline system, the positive bars indicate the percentage of intents (over those with count < $\tau$) that exceed 98\% than the system without augmentation. Models pcVAE and T5\_5x have the best performance, while cVAE deteriorates the original replication accuracy performance. See Appendix \ref{sec:app_extrinsic} for additional results. 

Overall we make the following key observations:
\begin{enumerate}[leftmargin=*]
    \vspace{-7pt}
    \item Text diversity seems to weigh more than text quality to boost replication accuracy, while simply oversampling does not provide much gain.
    \vspace{-19pt}
    \item Increasing augmentation size is more useful when generated data quality is high intrinsically.
    \vspace{-19pt}
    \item Data augmentation is more useful for intents that have lower counts.
\end{enumerate}


\section{Conclusion}

In this paper, we investigated the use of heterogeneous data augmentation of enhancing the skill routing system robustness for tail data. We empirically explored VAE-based, conditional BERT and T5-based structures to jointly model the routing hypothesis distribution with heterogeneous fields. In addition to intrinsic evaluation, we also we showed  that applying such data augmentation can be an effective way to increase replication performance on tail requests on a real-world dataset. 

\section*{Limitations}
This work proposes augmentation of heterogeneous data through various generative models in order to improve skill routing accuracy over tail traffic segments. The main challenge of such scenaria is two fold, namely generating meaningful high-quality text, as well as maintaining the joint distribution of the natural dataset across all present heterogeneous fields in the artificial set. In our experiments, we have carried out tests over real data, and multiple fields with varying types have been modeled here. In practice there could be many more fields that are present in the data, and expanding this framework over more fields could be beneficial. However, as the dimension and heterogeneity of the input fields increase, the task of learning such high-dimensional joint distribution grows harder, which is necessary in order to generate sound data where various realization of these fields in a given sample  are meaningful. More tests on such expanded setup may reveal that extra measures for promoting such behaviour is required during training.

\section*{Ethics Statement}
This paper is base on utilizing generative language models in order to increase the size of the training data through artificially-generate  real-looking data samples towards improving replication of the successful requests in tail traffic segments. Although we do not see direct ethical risks regarding this approach, one possible aspect could be that there is no direct evaluation of the generated data with respect to containing any sensitive samples such as profanity or discrimination. Nevertheless, given clean  training data for training such generative models, samples generated via these models are also expected to be free of such cases. We did not observe any inappropriate samples in the artificially generated texts, and performing data cleansing for the training set is out of scope for this work. 



\bibliography{anthology,custom}
\bibliographystyle{acl_natbib}

\newpage

\appendix

\section{A closer review on relevant prior work}
\label{sec:appendix_related_work}

\noindent\textbf{Data Augmentation in Task-oriented Dialogs.} To improve the model robustness in task-oriented dialogs \citep{chen_review21}, surface modification on texts like paraphrasing \citep{gao-etal-2020-paraphrase, cho-etal-2019-paraphrase, Einolghozati19} or token replacement \citep{Louvan2020SimpleDA, cbert18} is popular to introduce additional synthetic data for training and evaluating. Latent space modification like introducing noises \citep{kurata16b_interspeech}, adversarial training \citep{Lee2021ContrastiveLW} and data mixing \citep{mixmatch} is another research line to enhance the smoothness of predicted data distribution yet with lower interpretability. Many systems also augment out-of-scope queries to cope with few-shot or zero-shot NLU scenarios \citep{Chen2021GOLDIO, oodGAN}. In skill routing, slice-aware structure \citep{wang21} and random noise injection \citep{li21} are two closest works to improve model robustness. We follow the research heuristics to propose more advanced data augmentation frameworks to directly apply large-scale modification on routing data.

\noindent\textbf{Conditional Text Generation.} Besides modification on original data, directly modeling conditional text distribution is another promising category for advanced data augmentation. Seq2seq models \citep{Hou2018SequencetoSequenceDA, peng20} are widely used to directly model output distribution. GAN-based approaches \citep{Golovneva2020GenerativeAN, zhou19} implicitly train a text generator that simulates real utterances, which may somehow suffer from the mode collapse issue. Conditional Variational Auto-Encoder (cVAE) models like modeling conditional samples \citep{yoo18, Duan2020PretrainAP, malandrakis-etal-2019-controlled}, introducing structure attention \citep{qiu-etal-2020-structured} or leveraging pretrained models \citep{Xia2020CGBERTCT} are also an important family to improve over seq2seq models for generating more diverse and relevant text. However, these works pay much attention on pure text generation where we extend such framework to incorporate augmentation possibility in the heterogeneous form of routing data.

\section{Details of Methods}
\label{sec:appendix_methods}

In this section we provide a detailed discussion on our generative model adaptations based on conditional VAEs and BERT. 

\subsection{Conditional VAE (cVAE) with prior network (pcVAE)}
\label{sec:cvae}

We first consider an adaptation of Variational Autoencoder (VAE) for conditional text generation in Figure \ref{fig:vae_model}, which is one of the most popular frameworks for text generation. 
The basic idea of VAE is to encode the input text $x$ into a latent code $z$. To control the generation, a decoder network then digests the concatenation of $z$ and condition embeddings $Emb(c)$ from the desired NLU interpretations $n_1$ and the skill $s_1$ to reconstruct the original input using samples from $z$. During inference, we only sample $z$ from the prior distribution $\mathcal{N}(0,I)$ for decoding.

However, such model only learns an invariant representation $z$ that is independent of the desired conditions. To better constrain the behavior of conditional randomness, we follow \citet{zhao17} to introduce a prior network that models the conditional prior distribution $P(z|c)$ and use another recognition network $q(z|x,c)$ to approximate the true posterior distribution $P(z|x,c)$. The variational lower bound can be written as:
\begin{align}
    \notag
    \mathcal{L}(x,c)&=E_{q(z|x,c)}[log P(x|z,c)] \\
    \notag
                    &-D_{KL}(q(z|x,c)||P(z|c)) \\
                    &\leq log P(x|c)
\end{align}
During inference, we first sample a latent variable $z$ from the conditional prior distribution $\mathcal{N} (z|\mu', {\sigma'}^2 I)$ from the prior network and generate texts through the decoder $P(x|z,c)$. We use GRU networks for both text encoder and decoder.

\subsection{Conditional Variational BERT with Mask Language Modeling (CV-BERT MLM)}
\label{sec:cvbert}

Without much training power, the generated texts from the above models could be rather arbitrary and not grammatically sound. Moreover, we would like to generate other data fields besides texts, such as device type and device status signal represented as categorical ids. The most obvious way is to randomly perturb them, which nevertheless renders undesirable samples that may not follow the original joint distribution.
To ensure the quality and diversity of generated samples, we propose a new conditional augmentation framework in Figure \ref{fig:main} by incorporating cVAE in the backbone of BERT. 

Formally, we dissect BERT into two parts, where the first 6 layers serve as the encoder and the last 6 become the decoder. The input sequence is the concatenation of categorical fields and the utterance text. Suppose we have 3 categorical fields: $(C_1, C_2, C_3)$ e.g. (\texttt{Device\_id}, \texttt{Source\_id}, \texttt{Target\_id}) and a utterance $x=(W_1,...,W_T)$, giving the following input serquence:
\begin{align}
    (C_1, C_2, C_3, \texttt{[CLS]}, W_1,...,W_T, \texttt{[SEP]})
\end{align}
We then follow the mask language modeling (MLM) task in BERT to mask a few random tokens in the input sequence and aim to recover these masked tokens at the BERT output during training.

However, our goal is not to replicate the same tokens, instead replacing them with reasonably sound counterparts to augment the original data. 
Therefore, we incorporate cVAE idea mentioned in Section \ref{sec:cvae} inside BERT.
We use the encoder to model the true posterior distribution $P(z|x,c) \sim \mathcal{N}(z|\mu,\sigma^2 I)$ and minimize its KL divergence with another conditional prior distribution $P(z|c) \sim \mathcal{N}(z|\mu',\sigma'^2 I)$.
Here we replace the embedding of \texttt{[CLS]} token from the encoder output with the sampled $z$ to create the variation. The decoder will digest the perturbed encoder outputs to reconstruct the missing tokens. 
To preserve the semantics of the original sentence, we replace the segment embeddings in BERT with the embeddings $Emb(c)$ from the desired NLU interpretation and skill and attach it to each time step to control the text generation.

After training, we first perform an additional part-of-speech tagging step on the input sequence to be perturbed and determine only verbs and nouns for masking, which are more meaningful to alter. Then we sample a latent variable $z$ from the trained prior network $P(z|c)$ to insert in BERT and generate a new perturbed sequence.


\section{Experimental setup}
\label{sec:app_experiment_setup}

For cVAE and pcVAE, we set the word embedding size as 512 and the utterance encoder has the hidden size of 1024 for each direction. The context encoder has a hidden size of 128 and the response decoder has a hidden size
of 1024. The prior network consists of two fully connected networks with bottleneck hidden size as 100 and tanh non-linearity. The latent variable $z$ has a size of 128. The batch size is set as 256 and the models are trained end-to-end using the Adam optimizer with a learning rate of 0.001 and 30 epochs. We set the fixed KL loss weight as 0.1. For CV-BERT MLM, we adopt the pretrained $\mathbf{BERT}_{base}$ as the main backbone model and include the same prior network with $z$ size as 768. We set the mask probability of 90\% for categorical fields and 30\% of utterances. We train BERT models for 15 epochs with a learning rate of 5e-5. For T5, we set the batch size as 16, learning rate as 1e-4 and training epochs as 5. We run each experiments on 6 Tesla V100 GPUs.

\section{Intrinsic Evaluation Metrics}
\label{sec:app_intrinsic_metric}

 We use an ensemble of automatic metrics to evaluate the performance of generated samples without any expert-defined labels. For VAE-based methods, we evaluate the text performance in terms of reconstruction accuracy, fluency and diversity:
\begin{enumerate}[leftmargin=*] 
    \item \textbf{Reconstruction accuracy}: we use test reconstruction loss, BLEU-1,2 between generated and original texts.
    \vspace{-7pt}
    \item \textbf{Fluency}: we use language model perplexity to evaluate how fluent the generated texts are.
    \vspace{-7pt}
    \item \textbf{Diversity}: we use the unique rate as the proportion of unique sentences across all sentences.
\end{enumerate}
For all other methods, since there is no direct comparison in terms of reconstruction. We focus on the metrics of fluency and diversity. We also introduce additional diversity metrics following \citet{zhang18m} to increase the granularity of diversity evaluation. We use \textbf{Dist-1} and \textbf{Dist-2} to calculate the 
percentage of the number of unique 1-grams/2-grams over all 1-grams/2-grams of tested sentence. Further, we calculate \textbf{Ent-4} as how evenly the empirical 4-gram distribution is for a given sentence ($F(w)$ is the frequency of word $w$).
\begin{equation}
    Ent=-\frac{1}{\sum_w F(w)} \sum_{w\in V} F(w)log \frac{F(w)}{\sum_w F(w)}
\end{equation}
To evaluate the quality of categorical field generation whether it follows yet not reconstructs the original distribution, we perform the monte carlo sampling on predicted distribution from generative models and calculate the KL divergence (\textbf{Field KL}) with the ground truth distribution. 

\section{Intrinsic Evaluation Results} \label{sec:app_intrinsic_eval}

Table \ref{table:vae}  shows a comparison of results for two VAE methods in different target domains. First, we observe that larger training data will render better generalization power, which gives better reconstruction accuracy and lower perplexity in all four domains. Individual domains have rather disparate statistics, where the knowledge domain has the poorest performance with a high variance of utterance distribution. By introducing a prior network, it provides a better perplexity and diversity which conditions on top-1 NLU interpretations more explicitly.

\begin{table*}[t]
\footnotesize
\centering 
\resizebox{0.8\linewidth}{!}{
\begin{tabular}{l|l||c|c|c|c|c|c|c|c|c|c}
\hline
\multicolumn{2}{c||}{\textbf{Metrics}} & \multicolumn{6}{c|}{\textbf{Reconstruction accuracy}} & \multicolumn{2}{c|}{\textbf{Fluency}} & \multicolumn{2}{c}{\textbf{Diversity}} \\
\hline
 & & \multicolumn{2}{c|}{\textbf{Test loss}} & \multicolumn{2}{c|}{\textbf{BLEU-1}} & \multicolumn{2}{c|}{\textbf{BLEU-2}} 
 & \multicolumn{2}{c|}{\textbf{Perplexity}} & \multicolumn{2}{c}{\textbf{Unique rate}} \\
\hline
\textbf{Model} & \textbf{Domain} & 1M & 4M & 1M & 4M & 1M & 4M & 1M & 4M & 1M & 4M \\
\hline\hline
cVAE  & Knowledge & 3.261	&	1.701	&	0.902	&	0.947	&	0.813	&	0.886	&	25.39	&	5.307	&	0.9110	&	0.8928 \\
cVAE  & Shopping  & 2.019	&	1.184	&	0.928	&	0.956	&	0.666	&	0.710	&	7.385	&	3.190	&	0.8557	&	0.8680 \\
cVAE  & Video     & 1.789	&	1.059	&	0.933	&	0.957	&	0.767	&	0.809	&	5.843	&	2.804	&	0.8891	&	0.8984 \\
cVAE  & Books     & 2.016	&	1.299	&	0.943	&	0.958	&	0.853	&	0.881	&	7.365	&	3.579	&	0.8241	&	0.8508 \\
\hline
pcVAE  & Knowledge & 1.775	&	0.675	&	0.941	&	0.974	&	0.875	&	0.931	&	5.686	&	1.968	&	0.9596	&	0.9188 \\
pcVAE  & Shopping  & 1.021	&	0.643	&	0.960	&	0.970	&	0.719	&	0.737	&	2.704	&	1.905	&	0.6767	&	0.7195 \\
pcVAE  & Video     & 0.981	&	0.625	&	0.960	&	0.968	&	0.817	&	0.827	&	2.587	&	1.869	&	0.8321	&	0.8727 \\
pcVAE  & Books     & 1.185	&	0.749	&	0.963	&	0.969	&	0.890	&	0.899	&	3.186	&	2.120	&	0.8302	&	0.9021 \\
\hline
\end{tabular}
}
\caption{Intrinsic evaluation for VAE-based methods. 1M and 4M indicate the number of data for training the models. } 
\label{table:vae} 
\end{table*}

\section{Extrinsic Evaluation Results}
\label{sec:app_extrinsic}

\begin{figure}[t]
  \centering
  \includegraphics[width=1\linewidth]{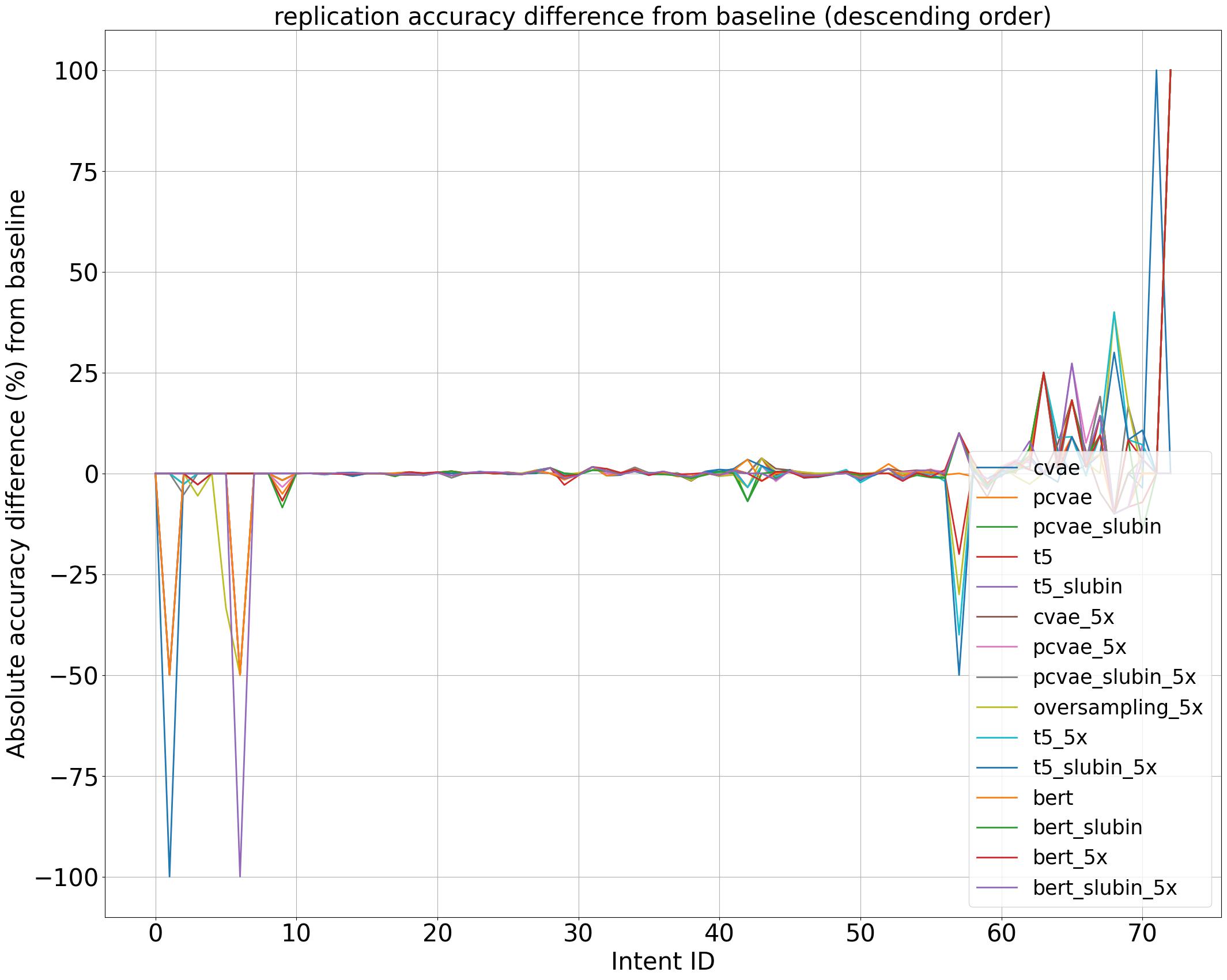}
  \caption{Replication accuracy difference from the baseline. We sort the intent based on their accuracy in the descending order and show the accuracy difference from the baseline.}
  \label{fig:replication_acc}
\end{figure}

\noindent\textbf{Sorting the intents in terms of accuracy}:
To see the improvement/regression from each intent pair, in Figure \ref{fig:replication_acc} we first sort the raw replication accuracy with the descending order and calculate the difference of the accuracy between each method and the baseline.

\begin{figure}[t]
  \centering
  \includegraphics[width=1\linewidth]{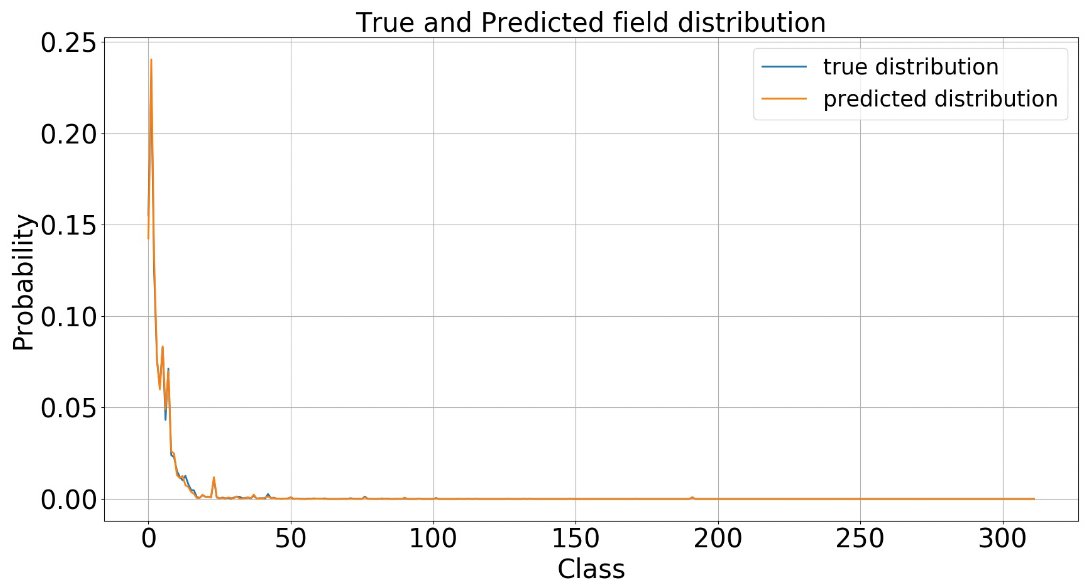}
  \caption{True and Predicted field distribution of device id comparison.}
  \label{fig:kl}
\end{figure}

Furthermore, Figure \ref{fig:kl} shows the sampling predicted and true distribution of device id from the best T5 model. We can see the model nicely follows the original distribution with its strong representation power.

\end{document}